\documentclass[a4paper, 10pt, conference]{ieeeconf}      
\usepackage[left=54pt, right=37pt, top=106pt, bottom=56.5pt]{geometry}
\IEEEoverridecommandlockouts                              

\usepackage{graphicx}
\usepackage{subcaption}
\usepackage{color}
\usepackage{amssymb}

\title{\huge \bf
Model-free Deep Reinforcement Learning for Urban Autonomous Driving}

\author{Jianyu Chen*, Bodi Yuan* and Masayoshi Tomizuka
\thanks{* indicates equal contribution}
\thanks{J. Chen, B. Yuan and M. Tomizuka are with Department of Mechanical Engineering, University of California, Berkeley, CA94720, USA. \quad Corresponding to {\tt\small jianyuchen@berkeley.edu}}%
\thanks{This work was supported by Denso International at America}
}

\begin{document}

\maketitle
\thispagestyle{empty}
\pagestyle{empty}

\begin{abstract}
Urban autonomous driving decision making is challenging due to complex road geometry and multi-agent interactions. Current decision making methods are mostly manually designing the driving policy, which might result in sub-optimal solutions and is expensive to develop, generalize and maintain at scale. On the other hand, with reinforcement learning (RL), a policy can be learned and improved automatically without any manual designs. However, current RL methods generally do not work well on complex urban scenarios. In this paper, we propose a framework to enable model-free deep reinforcement learning in challenging urban autonomous driving scenarios. We design a specific input representation and use visual encoding to capture the low-dimensional latent states. Several state-of-the-art model-free deep RL algorithms are implemented into our framework, with several tricks to improve their performance. We evaluate our method in a challenging roundabout task with dense surrounding vehicles in a high-definition driving simulator. The result shows that our method can solve the task well and is significantly better than the baseline.


\end{abstract}

\section{INTRODUCTION}
A highly intelligent decision making system is crucial for urban autonomous driving with dense surrounding dynamic objects. It must be able to handle the complex road geometry and topology, complex multi-agent interactions, and accurately follow the high level command such as routing information.

Current autonomous driving decision making systems are focusing on the non-learning model-based approaches, which often requires to manually design a driving policy~\cite{gonzalez2016review,paden2016survey}. Although it might not be difficult to design a qualified driving policy with the help of human prior knowledge, the manually designed policy could suffer from several weaknesses: 1) Accuracy: the predefined motion heuristics and models will lead to bias, especially for highly interactive environments. 2) Generality: for different scenarios and tasks, the model might need to be redesigned manually.

Recent advances in machine learning enables the possibility for learning based approaches for autonomous driving decision making. The most popular approach is imitation learning, which can learn a driving policy automatically from expert driving data~\cite{bojarski2016end,codevilla2018end,bansal2018chauffeurnet,chen2019deep}. However, there are some shortcomings for imitation learning: 1) It needs to collect a huge amount of expert driving data in real-world and in real-time, which can be costly and time consuming. 2) It can only learn driving skills that are demonstrated in the dataset. This might lead to serious safety issues because expert drivers generally do not provide dangerous demonstrations so the autonomous vehicle cannot learn how to deal with those dangerous cases. 3) Since the human driver experts act as the supervision for learning, it is impossible for an imitation learning policy to exceed human-level performance. 

Combined with deep learning techniques, reinforcement learning (RL) has brought a series of breakthroughs in recent years. Agents trained with deep reinforcement learning achieves super-human-level performance in Atari game playing~\cite{mnih2013playing,mnih2015human}, go playing~\cite{silver2016mastering,silver2017mastering}, and complex strategic games~\cite{vinyals2019alphastar,arulkumaran2019alphastar}.  With reinforcement learning, a policy can be learned automatically without any expert data. It can simulate various kinds of different cases, including some dangerous ones. It is also possible to achieve better performance than a human expert. Thus a good alternative to imitation learning for autonomous driving decision making is to use deep reinforcement learning.

However, there are not many successful applications for deep reinforcement learning in autonomous driving, especially in complex urban driving scenarios. This is due to: 1) Most of the methods directly use front view image as the input and learn the policy end-to-end. The extremely complex high dimensional visual features dramatically enlarge the sample complexity for learning. 2) Deep reinforcement learning is a fast evolving research area, but its application to autonomous driving has lag behind. Most researchers are still using basic deep RL algorithms such as deep Q network, which is not able to solve some complex problems. Much more powerful deep RL algorithms were developed in recent years but few of them have been applied to autonomous driving tasks.

In this paper, we design a specific input representation to reduce the sample complexity instead of directly using the front view image. We then use visual encoding to capture the low-dimensional latent states for urban autonomous driving tasks, which makes the problem more tractable for reinforcement learning. Several state-of-the-art model-free deep RL algorithms are implemented to learn a driving policy in a complex roundabout scenario with multiple surrounding vehicles running. Several tricks are developed to improve the performance of the algorithms, including modified exploration strategies, frame skip, network architectures and reward designs. Final result shows that our method can robustly learn a driving policy that is able to navigate through the complex urban driving scenario.

\section{RELATED WORKS}
With recent progresses in deep learning, learning-based approaches find their applications in autonomous driving, both in real world and in simulation. NVIDIA used a deep convolutional neural network to learn a lane following policy end-to-end from the font view image~\cite{bojarski2016end,bojarski2017explaining}. Waymo also used imitation learning to learn a urban driving policy from a huge amount of human driver data~\cite{bansal2018chauffeurnet}. Based on CARLA, an open-source simulator for autonomous driving research, ~\cite{codevilla2018end,chen2019deep} applied deep imitation learning to learn a policy to navigate through a complex virtual urban environment.

Researchers have also tried reinforcement learning approaches for autonomous driving. Wolf et al.~\cite{wolf2017learning} used a Deep Q Network to learn to steer an autonomous car in simulation. The action space is discrete and only allows coarse steering angles. Lillicrap et al.~\cite{lillicrap2015continuous} developed a continuous control deep reinforcement learning algorithm which is able to learn a deep neural policy to drive the car on a simulated racing track. Sallab et al.~\cite{sallab2017deep} proposed a deep RL framework for autonomous driving and applied it to the lane following task in simulation. Chen et al.~\cite{chen2018deep} proposed a hierarchical deep RL approach which is able to solve some complex temporal delayed problems such as traffic light passing. Nonetheless, the above approaches were developed either for simple scenarios without complex road geometry and multi-agent interaction, or used manually designed feature representations or policy models. 

Bird-view representation for autonomous driving decision making is proposed recently to reduce the complexity of visual features~\cite{djuric2018motion,bansal2018chauffeurnet,chen2019deep}. They generally compress information of the map and objects to a rasterized image. Direct RGB images are very high dimensional, one can use variational auto-encoder (VAE)~\cite{kingma2013auto,doersch2016tutorial} for dimension reduction, which learns a low dimensional latent representation of the image via an unsupervised style. 

Model-free deep reinforcement learning is a rich research area and is evolving rapidly. Related algorithms range from the Q learning based approaches such as DQN~\cite{mnih2013playing,mnih2015human} and double DQN~\cite{van2016deep}, the actor-critic approaches such as A3C~\cite{mnih2016asynchronous}, DDPG~\cite{lillicrap2015continuous} and TD3~\cite{fujimoto2018addressing}, the policy optimization approaches such as TRPO~\cite{schulman2015trust} and PPO~\cite{schulman2017proximal}, and the maximum entropy approaches such as soft actor-critic~\cite{haarnoja2018soft,haarnoja2018}.

\vspace{-0.5mm}

\section{PROBLEM FORMULATION}
Directly applying reinforcement learning algorithms on the raw sensor input can hardly work well on the urban autonomous driving problem. We thus proposed a framework to reduce the complexity of the problem, making it possible to be solved well by the current model-free deep reinforcement learning techniques. 

\begin{figure}
\centering
  \includegraphics[width = .45\textwidth]{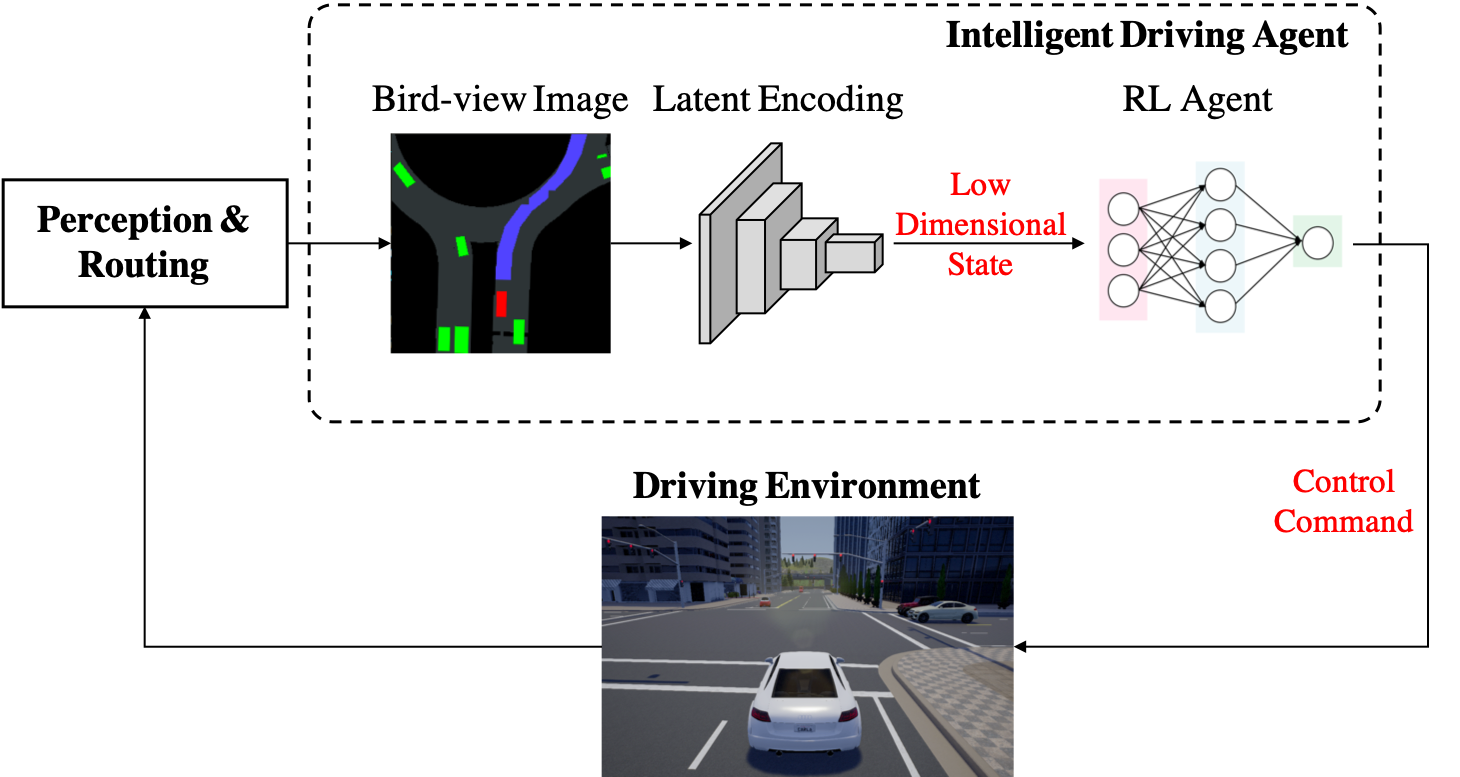}
  \caption{\label{Fig:framework}\em{Framework overview of the our system. The agent takes information from the perception and routing modules, generates a bird-view image and encodes it to low-dimensional latent states. Reinforcement learning is then applied to learn a policy to generate the correct control command.}}
  \vspace{-5mm}
\end{figure}

Fig.\ref{Fig:framework} shows the proposed framework. Our intelligent driving agent receives perception and routing information from the driving environment. It then processes the information into a bird-view image as the input representation. The image is then encoded into low dimensional latent states. Reinforcement learning algorithms are then adopted to train a deep neural policy which takes the encoded states as input and generates the control command such as acceleration and steering angle.

\subsection{Input Representation}
The input to the autonomous driving agent needs to cover enough information for urban driving scenarios. A straight forward input representation is to directly use the raw sensor data such as front view image. However, the raw sensor data contains extremely high dimensional information such as appearances and textures of the roads and objects, weather conditions, and light conditions. In order to obtain good generalization, the dataset must cover enough data for each dimension of the raw sensor information.

With the help of a perception module, the above complexity can be reduced largely. For example, we can use object detection to get the position, heading and velocity of each surrounding object; we can obtain lane related features such as longitudinal and lateral distance to a specific lane marking; we can also represent high level routing information as a set of trajectory points. However, although this kind of representation has been widely used with model-based policies, it brings problems for learning-based policies: 1) The number of surrounding objects is varying; 2) The information about road geometry and topology is highly structured. These information is extremely difficult to be transformed to a tensor with fixed shape, which can be used as the input to a learning based neural network policy.

\begin{figure}
    \centering
    \includegraphics[width = .45\textwidth]{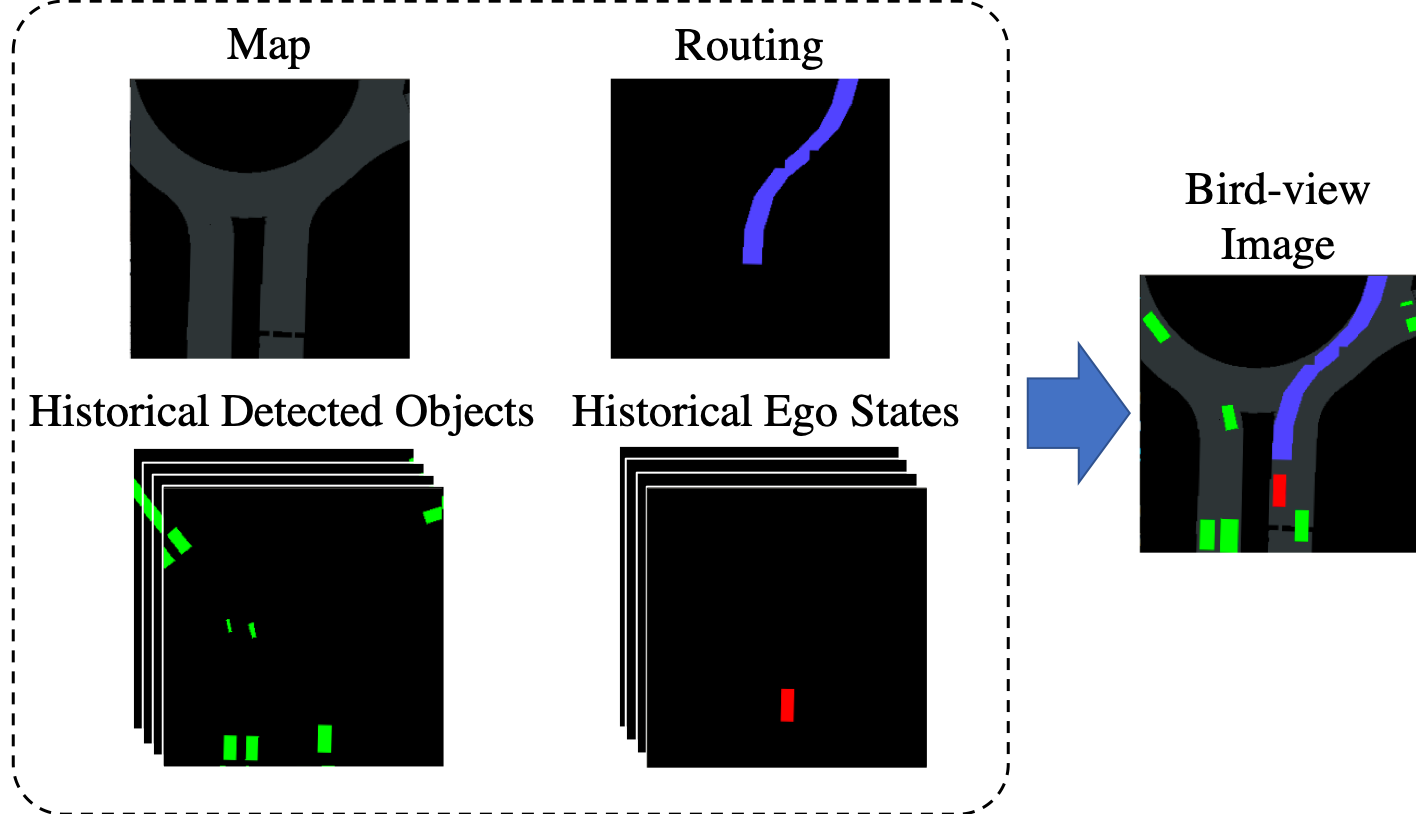}
    \caption{\em{Input Representation of our framework. The bird-view observation combines information of map, routing, historical detected objects and historical ego states.}}
    \label{Fig:representation}
    \vspace{-6mm}
\end{figure}

To alleviate this problem, we proposed to convert the output of perception module (object detection and localization), as well as the routing information, to a bird-view representation that applied as the input to our policy. This bird-view representation contains enough information for decision making, and is a big simplification of the complex visual and spatial information of raw sensor data. As shown in Fig.\ref{Fig:representation}, our bird-view input representation is composed of the following four parts:

\subsubsection{\bf{Map}}
The map contains information of road geometry. Here we render the drivable roads as gray polygons and other undrivable area part to be black.

\subsubsection{\bf{Routing}}
The routing information is calculated by a route planner. It contains information of a sequence of way points that the ego vehicle should follow. It is rendered as a thick blue polyline.

\subsubsection{\bf{Historical Detected Objects}}
The historical bounding boxes of detected surrounding objects (e.g, vehicles, bicycles, pedestrians) in a past sliding window are rendered as green boxes, with reduced level of brightness meaning earlier time-steps.

\subsubsection{\bf{Historical Ego States}}
Similar to the detected objects, the historical ego states are represented as boxes with reduced brightness. The color of the boxes are set red. 

\vspace{2mm}

The final bird-view image is rendered with pixel size $256 \times 256$ and resized to $64 \times 64$, and is always aligned with the ego vehicle view. The actual size of the field of view is $\left( {40m,40m} \right)$, where the ego vehicle is positioned at $\left( {20m,8m} \right)$.

\subsection{Latent State Encoding}
Even after we replace the complex raw sensor information with processed information, the input is still a high dimensional bird-view image. It will be extremely hard to learn a good policy with this high dimensional input. Meanwhile, the complex input can make it easy to get over-fitted during reinforcement learning process.

\begin{figure}
\centering
  \includegraphics[width = .47\textwidth]{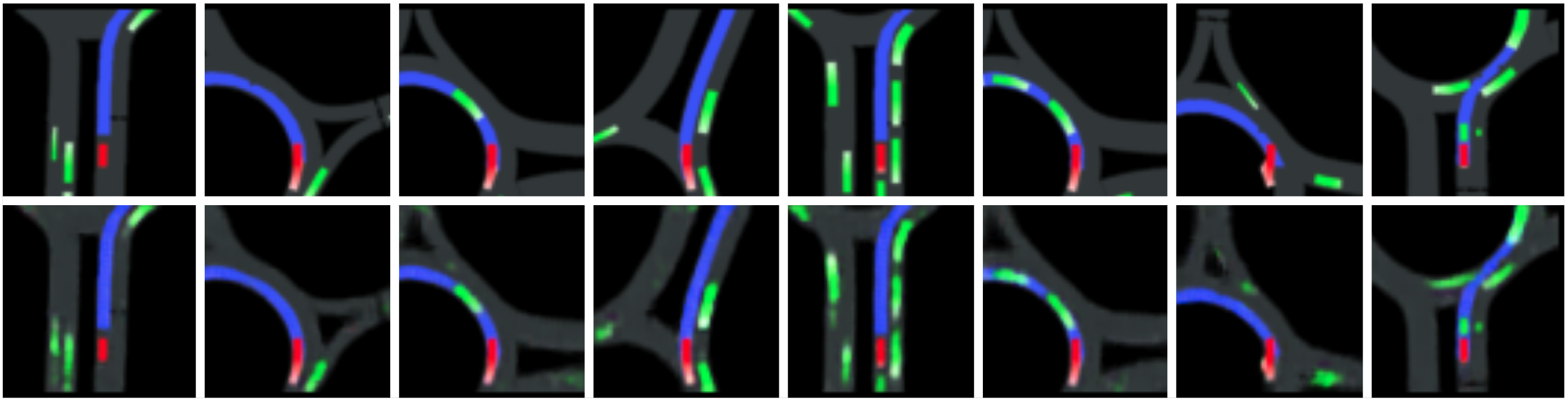}
  \caption{\label{Fig:vae}\em{Examples of the latent state encoding. The first row shows the original input images. The second row shows the reconstructed images based on the encoded latent states.}}
  \vspace{-5mm}
\end{figure}

In order to further reduce the input complexity, we propose to use variational auto-encoder (VAE)~\cite{kingma2013auto} to learn a low dimensional latent representation. VAE is composed of two parts, an encoding network  ${q_\phi }\left( {\left. {\bf{x}} \right|{\bf{o}}} \right)$ which encodes the original high dimensional observation $\bf{o}$ to a low dimensional latent state $\bf{x}$, and a decoding network ${p_\theta }\left( {\left. {\bf{o}} \right|{\bf{x}}} \right)$ which decodes $\bf{x}$ to $\bf{o}$. Here $\phi$ and $\theta$ denote the parameters of the encoder and the decoder, respectively. To obtain the parameters, the following objective needs to be maximized:
\vspace{-3mm}

{\footnotesize
\begin{equation}
{\cal L\left( {\phi ,\theta ,{\bf{o}}} \right)}
 = {D_{KL}}\left( {\left. {{q_\phi }\left( {\left. {\bf{s}} \right|{\bf{o}}} \right)} \right\| {p_\theta }\left( {\bf{s}} \right)} \right) - {\mathbb{E}_{{q_\phi }\left( {\left. {\bf{s}} \right|{\bf{o}}} \right)}}\left( {\log{p_\theta }\left( {\left. {\bf{o}} \right|{\bf{s}}} \right)} \right) 
 \end{equation}
 }
where $D_{KL}$ represents kullback-leibler (KL) divergence. The prior distribution is usually a multivariate Gaussian ${p_{\theta }(\mathbf {s} )={\mathcal {N}}(\mathbf {0,I} )}$. Some samples of the encoding results are shown in Fig.\ref{Fig:vae} where the first row shows the original input images, and the second row shows the reconstructed images. We can see the reconstructed images are pretty close to the original images, both for the road geometry and objects. This indicates that the encoded low dimensional latent state $\mathbf{s}$ successfully preserves the core information of the original image input.

\subsection{Reinforcement Learning}
With access to the state $s$ and reward function $r$, the objective of reinforcement learning is to find the optimal policy ${\pi ^ * }$ which optimize the expected future total rewards:
\begin{equation}
R\left( {\pi ,r} \right) = {\mathbb{E}_\pi }\left[ {\sum\limits_{t = 0} {{\gamma ^t}r\left( {{s_t},{a_t}} \right)} } \right]
\end{equation}
where $\gamma$ is the discount factor of the reward. The solution ${\pi ^ * }\left( {\left. {{a_t}} \right|{s_t}} \right)$ can then be used as the controller of our agent, which takes the current input state $s_t$ and output the control command $a_t$ to be applied to the vehicle. The next section will introduce how we can obtain the policy.

\section{ALGORITHMS}
Three state-of-the-art model-free deep reinforcement learning algorithms are applied in our framework to learn the driving policy. We will introduce them briefly in this chapter.
\subsection{Double Deep Q-Network (DDQN)}
The most representative model-free reinforcement learning is Q learning~\cite{watkins1989learning}, which is based on an estimation of the Q value ${Q_\pi }\left( {s,a} \right)$, defined as the expected future total rewards when taking action $a$ at state $s$ and then follow the policy $\pi$ thereafter:
\begin{equation}
{Q_\pi }\left( {s,a} \right) = {\mathbb{E}_\pi }\left[ {\left. {\sum\limits_{t = 0} {{\gamma ^t}r\left( {{s_t},{a_t}} \right)} } \right|{s_0} = s,{a_0} = a} \right]
\end{equation}

The optimal Q value is ${Q^*}\left( {s,a} \right) = \mathop {\max }\limits_\pi  {Q_\pi }\left( {s,a} \right)$ and the optimal policy is the highest valued action ${\pi ^ * }\left( s \right) = \mathop {\arg \max }\limits_a {Q^*}\left( {s,a} \right)$. Note that the action set for Q learning must be finite discrete. The optimal Q value can be learned using temporal difference (TD) learning~\cite{sutton1998introduction}.

When using a deep neural network to parametrize the Q value as $Q\left( {s,a;\theta } \right)$, where $\theta$ are parameters of the neural network, the state can be extended to high dimensional and continuous. The corresponding algorithm is called deep Q network (DQN)~\cite{mnih2013playing,mnih2015human}. DQN builds a replay buffer $\cal D$, and defines an additional target Q network with parameters ${\theta '}$, besides the online Q network with parameters $\theta$. During learning, the transition pairs $\left( {s,a,r,s'} \right)$ are stored into the replay buffer and uniformly sampled a mini-batch at each step. Then the learning targets are defined as:
\begin{equation}
y = r + \gamma \mathop {\max }\limits_{a'} Q\left( {s',a';\theta '} \right)
\label{Eq:dqn}
\end{equation}

Supervised learning is applied to minimize the objective $\mathop {\min }\limits_\theta  \sum {\left( {y - Q\left( {s,a;\theta } \right)} \right)}$ and $\theta$ is copied to ${\theta '}$ every $\tau $ steps.

The original DQN algorithm has the problem of overestimating the Q values. Thus Double DQN (DDQN)~\cite{van2016deep} is proposed to avoid the overoptimistic value estimates. The main idea is to separate action selection from the value estimation. DDQN thus obtains the action using the online Q network, but estimates the Q value using the target Q network. The algorithm is mostly same with DQN but replaces the learning target (\ref{Eq:dqn}) with:
\vspace{-1mm}
\begin{equation}
y = r + \gamma Q\left( {s',\mathop {\arg \max }\limits_{a'} \left( {s',a';\theta } \right);\theta '} \right)
\end{equation}

\subsection{Twin Delayed Deep Deterministic Policy Gradient (TD3)}
DQN and DDQN can only solve problems with finite discrete action spaces. However most environments need continuous actions. For continuous control, a policy network ${\pi _\phi }$ is introduced, and it can be optimized through the deterministic policy gradient algorithm (DPG)~\cite{silver2014deterministic}, known as actor-critic. The policy can be optimized by taking gradient steps with respect to the expected future total rewards:
\begin{equation}
{\nabla _\phi }J\left( \phi  \right) = {\mathbb{E}_\pi }\left[ {{{\left. {{\nabla _a}{Q_\pi }\left( {s,a} \right)} \right|}_{a = \pi \left( s \right)}}{\nabla _\phi }{\pi _\phi }\left( s \right)} \right]
\end{equation}

Combined with deep neural networks, the deep deterministic policy gradient (DDPG) algorithm~\cite{lillicrap2015continuous} is proposed. Similar to DQN algorithm, DDPG also builds both an online and a target Q network with parameters $\theta$ and $\theta '$, as well as an online and a target policy network with parameters $\phi$ and $\phi '$. Transitions are stored into a replay buffer and mini-batches are sampled at each step. To update the Q network, the Q value targets are set as:
\vspace{-2mm}
\begin{equation}
y = r + \gamma Q\left( {s',{\pi _{\phi '}}\left( {s'} \right);\theta '} \right)
\vspace{-1mm}
\end{equation}

The policy network is updated using the sampled policy gradient:
{\small
\begin{equation}
{\nabla _\phi }J\left( \phi  \right) = \frac{1}{N}\sum\limits_i {{{\left. {{\nabla _a}Q\left( {s,a;\theta } \right)} \right|}_{s = {s_i},a = {\pi _\phi }\left( {{s_i}} \right)}}{{\left. {{\nabla _\phi }{\pi _\phi }\left( s \right)} \right|}_{{s_i}}}} 
\end{equation}
\vspace{-2mm}
}

The target Q and policy networks are updated using temporal difference with respect to the online networks:  
\vspace{-1mm}
\begin{equation}
\begin{array}{l}
\theta ' \leftarrow \tau \theta  + \left( {1 - \tau } \right)\theta '\\
\phi ' \leftarrow \tau \phi  + \left( {1 - \tau } \right)\phi '
\end{array}
\vspace{-1mm}
\end{equation}

DDPG can suffer from the function approximation errors which lead to overestimated values and suboptimal policies. Thus the twin delayed deep deterministic policy gradient (TD3) algorithm~\cite{fujimoto2018addressing} is proposed to address the problem. It borrows the idea of double Q learning to build an additional Q network, and take the minimum value between the pair of Q networks when setting the target Q value. It also suggests to delay the target network update steps.

\subsection{Soft Actor Critic (SAC)}
Although the above methods work well in a range of challenging decision making tasks, they usually suffer from high sample complexity and brittle convergence property, which leads to extraordinary hyper-parameter tuning. The soft actor critic (SAC) algorithm~\cite{haarnoja2018soft,haarnoja2018} is proposed to alleviate this problem based on the theory of maximum entropy reinforcement learning. In this framework, the policy needs to maximize both the expected rewards and the entropy:
\vspace{-1mm}
\begin{equation}
\mathop {\max }\limits_\pi  J\left( \pi  \right) = \sum\limits_{t = 0} {{\mathbb{E}_\pi }\left[ {r\left( {{s_t},{a_t}} \right) + \alpha \mathcal{H} \left( {\pi \left( {\left.  \cdot  \right|{s_t}} \right)} \right)} \right]} 
\vspace{-1mm}
\end{equation}
where $\alpha$ is the weight of the entropy term. This modified reward function defines a so-called soft Q function and the corresponding soft Bellman backup operator:
\vspace{-1mm}
\begin{equation}
{\mathcal{T}^\pi }Q\left( {{s_t},{a_t}} \right) = r\left( {{s_t},{a_t}} \right) + \gamma {\mathbb{E}_{{s_{t + 1}}}}\left[ {V\left( {{s_{t + 1}}} \right)} \right]
\label{Eq:softQ}
\vspace{-2mm}
\end{equation}
where 
\vspace{-2mm}
\begin{equation}
V\left( {{s_{t + 1}}} \right) = {\mathbb{E}_{{a_t} \sim \pi }}\left[ {Q\left( {{s_t},{a_t}} \right) - \log \pi \left( {\left. {{a_t}} \right|{s_t}} \right)} \right]
\label{Eq:softV}
\vspace{-2mm}
\end{equation}
is the soft value function. It can be shown that applying the soft bellman backup will converge to the optimal soft Q value. SAC defines a soft Q network ${Q_\theta }$, a soft value network ${Q_\theta }$, and a policy network ${\pi _\phi }$, as well as an additional target value network ${V_{\psi '}}$. Similar to DQN and DDPG, the Q network and value network can be updated using supervised learning by defining the following targets according to (\ref{Eq:softQ}) and (\ref{Eq:softV}):
\vspace{-1mm}
\begin{equation}
\hat V\left( {{s_t}} \right) = {E_{{a_t} \sim {\pi _\phi }}}\left[ {{Q_\theta }\left( {{s_t},{a_t}} \right) - \log {\pi _\phi }\left( {\left. {{a_t}} \right|{s_t}} \right)} \right]
\vspace{-3mm}
\end{equation}
\vspace{-3mm}
\begin{equation}
\hat Q\left( {{s_t},{a_t}} \right) = r\left( {{s_t},{a_t}} \right) + \gamma {\mathbb{E}_{{s_{t + 1}}}}\left[ {{V_{\bar \psi }}\left( {{s_{t + 1}}} \right)} \right]
\vspace{-1mm}
\end{equation}

The policy is a stochastic neural network ${f_\phi }\left( {{\varepsilon _t};{s_t}} \right)$ where ${{\varepsilon _t}}$ is sampled from some fixed distribution such as a multivariate normal distribution. Then policy network can be optimized by applying policy gradient to the expected future rewards:
\vspace{-4mm}

{\footnotesize
\begin{equation}
{J_\pi }\left( \phi  \right) = {E_{{s_t},{\varepsilon _t}}}\left[ {\log {\pi _\phi }\left( {\left. {{f_\phi }\left( {{\varepsilon _t};{s_t}} \right)} \right|{s_t}} \right) - {Q_\theta }\left( {{s_t},{f_\phi }\left( {{\varepsilon _t};{s_t}} \right)} \right)} \right]
\end{equation}
}

\vspace{-6mm}

\section{EXPERIMENTS}
\subsection{Simulation Environment and scenario}
\begin{figure}
    \centering
    \includegraphics[width = .48\textwidth]{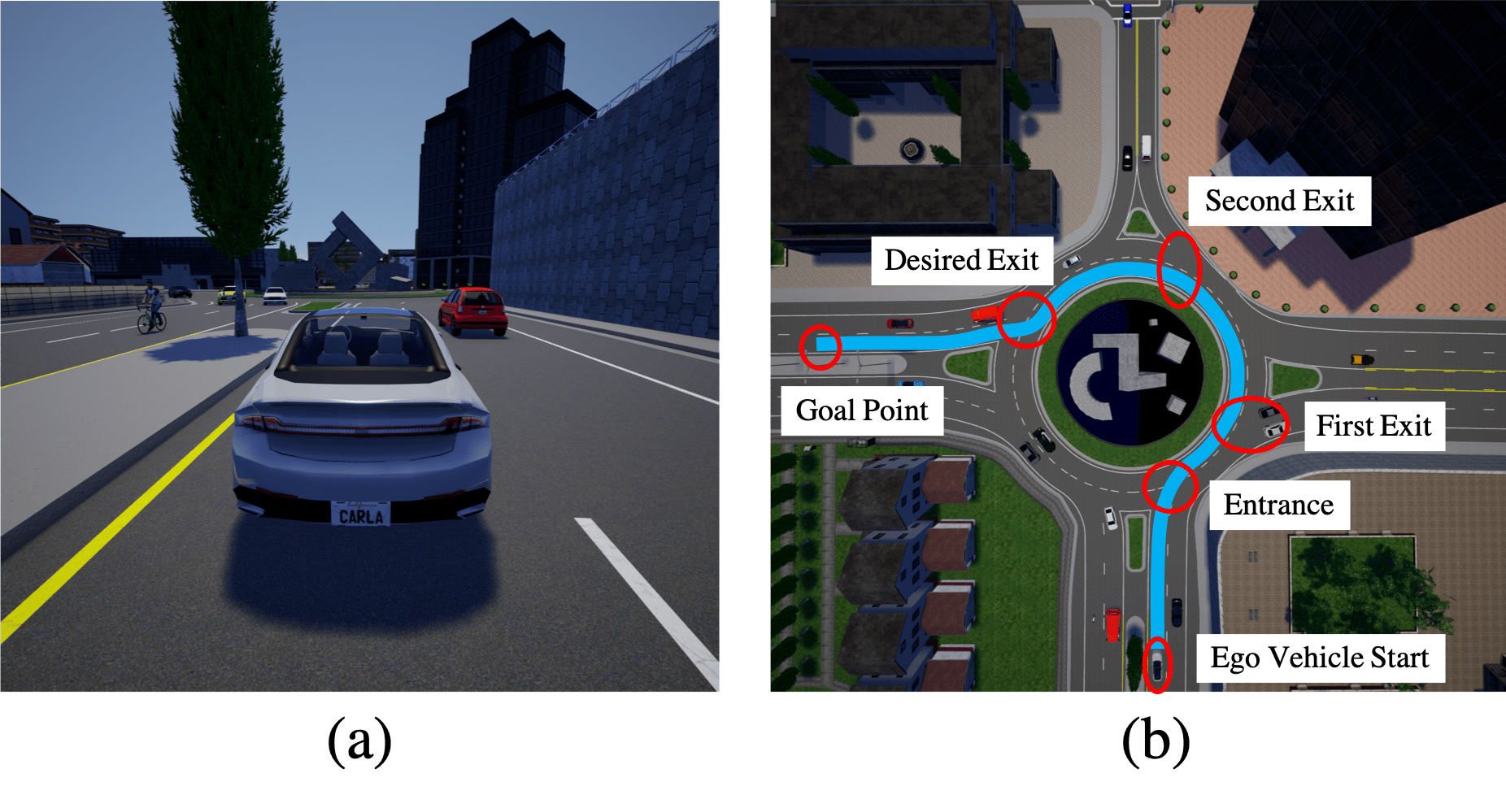}
    \caption{\em{The simulation environment and experiment scenario. Left is a sample view of the simulation. Right is a bird-view of our roundabout task scenario.}}
    \label{Fig:carla}
    \vspace{-7mm}
\end{figure}

We train and evaluate our method on the CARLA simulator~\cite{dosovitskiy2017carla}, which is a high-definition open-source simulator for autonomous driving research. Fig.\ref{Fig:carla} (a) shows a sample view of the driving simulation environment we use. It includes various urban scenarios such as intersection and roundabout. The range of the map is $400m \times 400m$, containing around $6km$ total length of roads. 

We choose the most challenging scenario, the central roundabout in this map for training and testing our method. As shown in Fig.\ref{Fig:carla} (b), the task is to start from an entrance of the roundabout, safely and efficiently enter the roundabout, pass through the first two exits, drive out to the desired exit and reach the final goal point. There are 100 vehicles initially sampled in the whole map, and nearly half of them are sampled around the roundabout, making the traffic quite busy. The sampled vehicles will randomly choose a direction when encountering with multiple choices of routes, then follow the route, and slow down if there are front vehicles.

\subsection{Network Architectures}
We train our variational auto-encoder (VAE) on 50k bird-view images, which is generated by using a simple controller to drive around the roundabout. Note that we do not need any labels in the training dataset, such as the action commands or ground truth positions of vehicles, but only the raw bird-view images. We also do not require the controller to drive well, it does not matter if the vehicle drives out of lane or collides with other objects. The VAE model has 4 conv-layers of $3\times 3$ kernel size, with 32, 64, 128, and 256 channels separately. The stride is set to 2. Each conv-layer is followed by ReLU activation. The latent space layer of size 64 is then fully connected to the last conv-layer. The VAE model is trained from scratch using Adam optimizer~\cite{kingma2014adam}, with learning rate $10^{-4}$ for 100 epochs. Some examples of trained VAE results are given in Figure~\ref{Fig:vae}.

During the reinforcement learning process, all networks have the same conv-layers and first dense layer copied from the pretrained VAE model, denoted as visual encoding layers here. To keep the visual encoding stable, the weights of the visual encoding layers are fixed without updating. Now we introduce the remaining layers of the networks used in the three reinforcement learning algorithms as follows:

\subsubsection{Double Deep Q-Network (DDQN)}
DDQN has one Q network and a target Q network with the same architecture. After the visual encoding layers, the network is then followed with 5 dense layers, with hidden layers ranging from 256 to 32 nodes. The output size is equal to the number of the possible actions, each representing the Q value of the corresponding action. The networks are trained using Adam optimizer with learning rate of $10^{-3}$.

\subsubsection{Twin Delayed Deep Deterministic Policy Gradient (TD3)}
TD3 has a policy network and two Q networks. After the visual encoding layers, all of these networks have 4 dense layers, with 64, 200 and 20 hidden units separately, followed by batch normalization and leaky ReLU for each layer. The negative slope of leaky ReLU is set to 0.01. The last dense layer of the policy network, corresponding to the action output, has 2 units, with a tanh activation function to limit the action range. The Q networks are a little different, their latent states from the visual encoding are first concatenated with the corresponding action, and the last layer has only one unit, denoting the Q value. Also, there is no batch normalization and no tanh activation used in Q networks. All the networks are trained using Adam optimizer. The policy network and Q networks are trained with learning rate of $10^{-4}$ and $10^{-3}$ respectively.

\subsubsection{Soft Actor Critic (SAC)}
The soft actor critic has two Q networks, one value network, and a policy network. The Q and value networks has the same architecture with one output unit. The visual encoding layers are followed by 5 dense layers with hidden units ranging from 256 to 32. The policy network's architecture is the same except that for the last layer, it splits to two branches. This is because the policy network of SAC is stochastic. The first branch represents the mean of the action and the second branch represents its variance. All networks are trained using Adam optimizer with learning rate of $3 \times 10^{-4}$.

\subsection{Implementation Details}
To improve the performance on the proposed task, we designed specific reward functions and added several tricks such as frame skip and new exploration strategies. 

\subsubsection{Rewards Design}
After testing a variety of reward combinations, we came up with a five-term reward function which works well. The reward function is given below:
\vspace{-2mm}
\begin{equation}
    r = r_{v} + r_{\alpha} + r_c + r_o + c .
\label{eq1}
\vspace{-2mm}
\end{equation}
where $r_{v}$ is the term encouraging moving forward to make progress. It is set equal to the ego vehicle's speed. Moreover, we set $r_{v} \gets 10-r_{v}$ if $r_{v} > 5$ to penalize exceeding speed limit. $r_{\alpha}$ is the term penalizing the magnitude of steering angle $\alpha$ to improve driving smoothness, where $r_{\alpha} = 0.5 * \alpha^2$. $r_c$ is the term penalizing collision with other surrounding vehicles, where $r_c = -10$ if there is a collision, otherwise $r_c = 0$. $r_o$ is the term penalizing running out of the lane, where $r_o = -1$ if the distance $d$ between the ego vehicle the routing baseline is larger than $2m$, otherwise $r_o = 0$ (note that we also tried to set $r_o$ continuously, such as linear to the distance $d$ or $d^2$, but it turns out the discrete one works better). The last term $c$ is a small constant set to $-0.1$, which was used to penalize the ego vehicle for stopping still.


\subsubsection{Frame Skip}
We use the frame skip trick during training, where we keep the action unchanged for $k$ consecutive frames. In other words, during the training, each action made by the ego vehicle will last $k$ frames until the new action starts to be effective. This technique heavily reduces the training complexity, one can regard it as reducing the search depth by a factor of $k$. 


However, $k$ shouldn't be set too large. If $k$ is too large, the correct action space might be too small or even not existed. For example, when we set $k=10$, it is difficult for the ego vehicle to make turns successfully. In our experiments, we use $k=4$.

\subsubsection{Exploration Strategies}


We proposed different exploration strategies in our implementation. For DDQN, we use a different kind of epsilon greedy strategy. Generally, epsilon greedy chooses the random actions from uniform distribution. However, the uniform distribution contains no heuristic information, which is inefficient. We thus propose to sample the random actions according to each action's Q values, where actions with bigger Q values will be sampled more frequently. This strategy introduces the information of expected future rewards and is thus more efficient. 

For TD3, we added a zero-mean Gaussian action noise with a specific variance $\sigma$:
\vspace{-2mm}
\begin{equation}
    \sigma = \delta * \lambda_{t} * \lambda_{d} * \lambda_{p},
\label{strategy}
\vspace{-2mm}
\end{equation}
where $\delta$ is set to $0.5$ and $0.1$ for acceleration and steering, respectively. $\lambda_{t}=\max(0.5, 1-t/T)$, where $t$ is the current training step and $T$ is the total decay time set to $100k$. $\lambda_{d} = \max(1-t/T, 0.2 + t_{p}/T_{p})$, where $t_{p}$ is the current path step in a single exploration path, and $T_{p}$ is the maximum path length which is set to $500$. We also introduce a periodic coefficient, $\lambda_{p} = 1+\sin(5\pi * t/T+\pi/2)$.


For SAC, the exploration strategy is actually included in the stochastic policy, which is learned and adapted during training, thus we do not need to add any specific exploration strategy for it.

\vspace{-2mm}
\section{RESULTS}
We evaluate our proposed framework using the metrics of average return, and success rate for entering and passing through the roundabout. The approaches are evaluated under two cases: without and with adding surrounding vehicles.

\subsection{Roundabout without Vehicles}
We first test our different RL approaches in the roundabout scenario without adding any surrounding vehicles. We also implemented a DDPG approach and trained directly on the front view images, with the same input sizes, network architectures, exploration strategies and hyper-parameters as in~\cite{lillicrap2015continuous}. This DDPG approach is set as the baseline.

\begin{figure}[!h]
\centering
  \includegraphics[width = .5\textwidth]{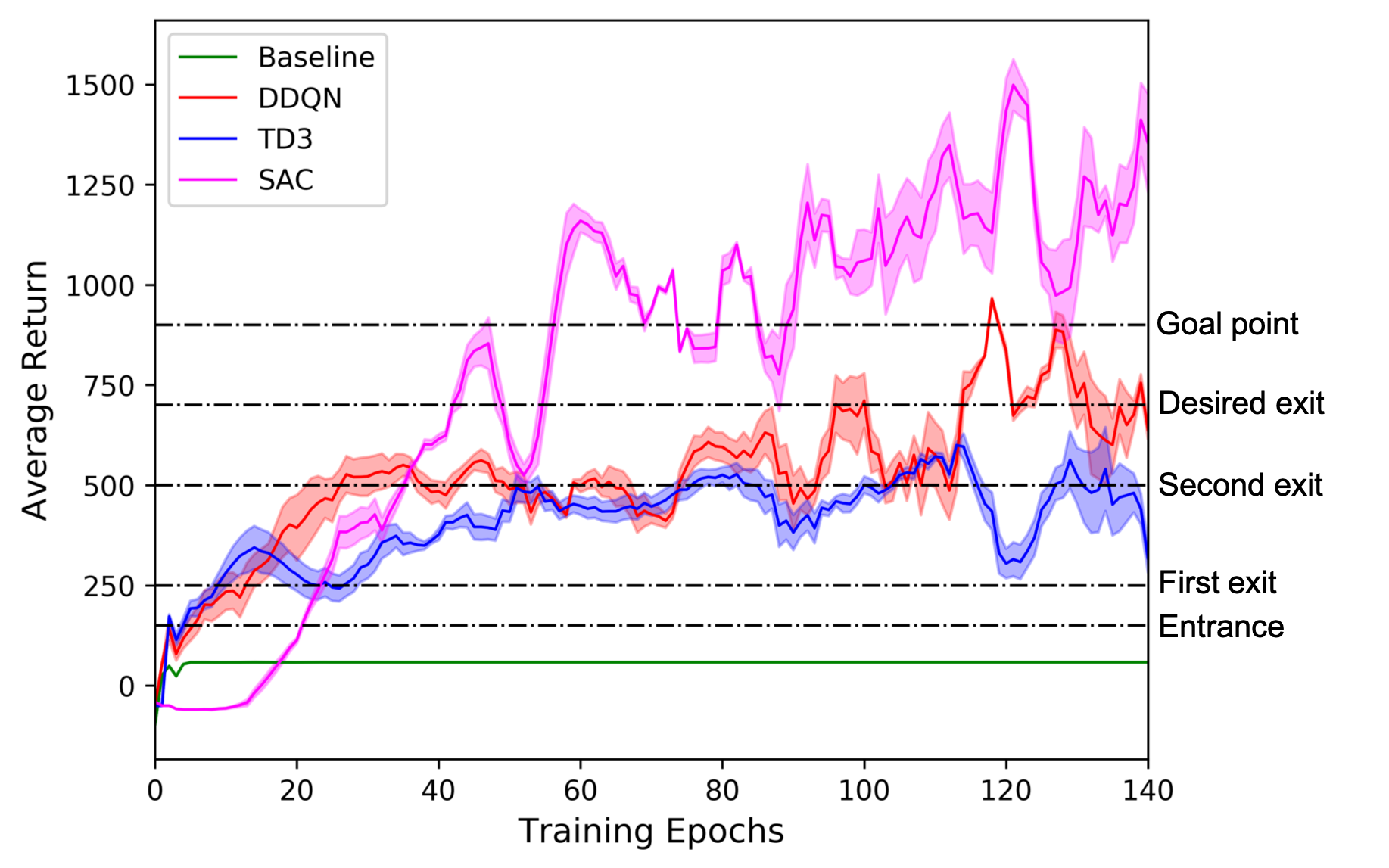}
  \caption{\em{Learning curves of different RL approaches applied in the roundabout scenario without adding vehicles. The shaded region represents half a standard deviation of returns with 1k evaluation steps. Curves are smoothed for visual clarity.}}
  \label{Fig:results_noCar}
  \vspace{-5mm}
\end{figure}

Fig.~\ref{Fig:results_noCar} shows the average returns for each approach. Rough estimations of returns for each checkpoint (entrance, first/second/desired exit, goal point) are drawn as horizontal dashed lines. As we can see, with the implementation under our proposed framework, SAC algorithm has the best performance, the trained ego vehicle can pass through the roundabout and reach the goal point (refer to Fig.~\ref{Fig:carla}) in most of times after 100 epochs. Although DDQN and TD3 make slow progress after the ego vehicle gets around the first two exits (refer to Fig.~\ref{Fig:carla}), all of the three algorithms perform well in the roundabout before the second exit. Due to our limited computation sources, we stopped at 140 epochs, when our SAC policy can already reach the final goal point stably. We believe that given longer training time and more computational resource, these algorithms can perform even better. On the other hand, the baseline has very low average returns and can barely learn anything. Actually, the vehicle just keeps turning right and goes out of the lane. The policy gets stuck in a local optimum and is not able to make any progress.

\subsection{Roundabout with Dense Surrounding Vehicles}

\begin{figure}[h]
\vspace{-4mm}
\centering
  \includegraphics[width = .5\textwidth]{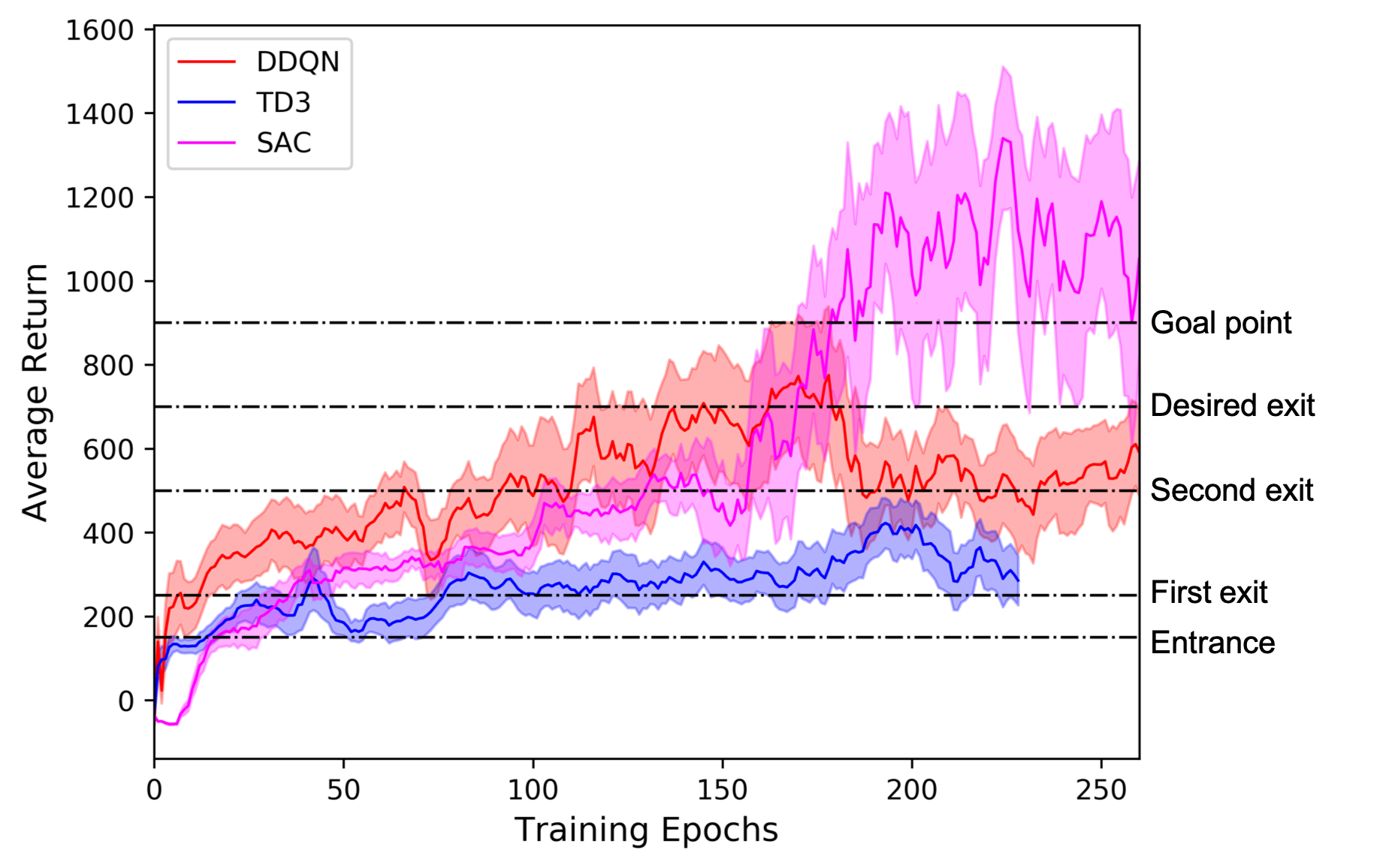}
  \caption{\em{Learning curves of different RL approaches for the roundabout scenario with adding 100 vehicles. The shaded region represents half a standard deviation of returns with 1k evaluation steps. Curves are smoothed for visual clarity.}}
  \label{Fig:results_100Cars}
  \vspace{-3mm}
\end{figure}

After adding dense surrounding vehicles to the environment, the task becomes much more complicated. In this scenario we only implement the three RL algorithms under our proposed framework but not the DDPG baseline since it can learn nothing even without any surrounding vehicles.

As shown in Fig.~\ref{Fig:results_100Cars}, SAC is still the best of all methods, and it can frequently reach the final goals after about 200 epochs. Model trained with SAC learns not only the lane following skills, but also how to interact with other vehicles appropriately in the busy roundabout. DDQN, TD3 can often successfully enter the roundabout or reach the first exit, but can barely reach the second exit. This might be due to the advantage of the adaptive exploration of SAC.



We then tested the three RL algorithms for 50 times each and summarized their success rates under our roundabout task with dense surrounding vehicles. The success rates are recorded at five checkpoints: the entrance, the first exit, the second exit, the desired exit, and the goal point (Fig.\ref{Fig:carla}). Results are shown in Table.~\ref{success}. We can see that the ego vehicle trained with all of the three approaches can successfully enter the roundabout with over 80\% success rate, indicating they have learned correct behaviors when entering the roundabout such as yielding to upcoming vehicles. When the distance increases, the success rate for DDQN and TD3 decreases dramatically. This is mainly because they are easy to suffer from insufficient exploration. Thanks to a good balance of exploration and exploitation, SAC performs the best and can reach the goal point quite often.

When we look at the failure cases, we found that almost all the failures come from rear-end to the front vehicle. This is because we do not explicitly incorporate velocity of vehicles in our input, which is essential for keeping safe distance with the front vehicle. Instead, we draw historical vehicle states using boxes with fading color, which only implicitly contain the velocity information. This problem is even more serious when the input is passed through an encoding network. As in Fig.\ref{Fig:vae}, we can barely see the fading color of some vehicles in the reconstructed images.

\begin{table}[!h]
\vspace{-1mm}
\centering
\caption{\em{Success rate for the roundabout scenarios evaluated on our three models DDQN, SAC and TD3. The value represent percentage of success trials.}}
{
\begin{tabular}{lccc}
\hline
Approach          & DDQN  & TD3  & SAC \\ \hline
Entrance          & 80\%  & 88\% &  86\%            \\
First exit    & 52\%  & 74\% &  80\%           \\ 
Second exit   & 38\%  &  2\% &  74\%           \\ 
Desired exit  &  8\%  &  0\% &  64\%           \\ 
Goal point        &  0\%  &  0\% &  58\%           \\ \hline
\end{tabular}}
\label{success}
\vspace{-4mm}
\end{table}

\section{CONCLUSIONS}
In this paper, we proposed a framework to enable model-free deep reinforcement learning in challenging urban autonomous driving scenarios. We designed a bird-view input representation to reduce the sample complexity, and used visual encoding to capture the low-dimensional latent states. We then applied three state-of-the-art model free deep RL algorithms (DDQN, TD3, SAC) into our framework, with several tricks to improve the performance. We evaluate our method in a challenging roundabout scenario with dense surrounding vehicles in a high-definition driving simulator. The results showed that that our method has the power to solve the tasks well. Although our method is significantly better than the baseline, it doesn't solve task perfectly. In the future, we will use more computation resources to dig the limit of our method, and also improve the learning efficiency. Furthermore, We will explore the generalization ability to other scenarios.

\vspace{-1mm}

\bibliographystyle{ieee}
\bibliography{reference}

\end{document}